\documentclass[11pt]{article}

\usepackage[letterpaper,margin=1in]{geometry}
\usepackage[T1]{fontenc}
\usepackage{mathptmx}
\usepackage[protrusion=true,expansion=false]{microtype}
\usepackage{amsmath,amssymb}
\usepackage{graphicx}
\usepackage{booktabs}
\usepackage{array}
\usepackage{siunitx}
\usepackage{float}
\usepackage[numbers,square,comma]{natbib}
\usepackage[hidelinks]{hyperref}
\usepackage{xcolor}
\usepackage[font=small,labelfont=bf,labelsep=colon]{caption}
\usepackage{subcaption}
\usepackage{fancyhdr}
\usepackage{enumitem}
\usepackage{titlesec}
\usepackage[htt]{hyphenat}

\titleformat{\section}{\Large\bfseries}{\thesection}{0.8em}{}
\titleformat{\subsection}{\large\bfseries}{\thesubsection}{0.7em}{}
\titleformat{\subsubsection}{\normalsize\bfseries}{\thesubsubsection}{0.6em}{}
\titlespacing*{\section}{0pt}{1.6em}{0.7em}

\newcommand{\ece}{\ensuremath{\mathrm{ECE}}}
\newcommand{\vit}{ViT-B/16}
\newcommand{\rn}{ResNet-50}

\hypersetup{
  pdftitle={Uncertainty-Aware Brain Tumor MRI Classification},
  pdfauthor={Author list},
}

\begin{document}

\thispagestyle{empty}
\begin{center}
  \vspace*{0.6in}

  {\huge\bfseries Monte Carlo Dropout Uncertainty and\\[0.45em]
   Entropy-Thresholded Selective Prediction for\\[0.45em]
   Architecture-Agnostic Brain Tumor MRI Triage\par}

  \vspace{2.4em}
  {\Large\bfseries Calibrated confidence for reliable defer-to-human decisions\par}

  \vspace{2.4em}
  {\Large\itshape A five-seed study of ViT-B/16 and ResNet-50 with temperature scaling and\\[0.25em]
   entropy-ranked selective prediction\par}

  \vspace{2.4em}
  {\large Medhansh Sharma\par}

\end{center}
\clearpage

\thispagestyle{fancy}
\begin{center}\section*{Abstract}\end{center}
\noindent
Deep networks now subtype brain tumors on MRI about as well as specialist readers, yet accuracy is
not what keeps them out of the clinic. What matters at the point of care is whether a model's
confidence can be trusted to flag the cases it is likely to misclassify and defer them to a human.
Deterministic estimates cannot: an auxiliary confidence head trained alongside the classifier
collapses to a near-constant output that says nothing about correctness. This study proposes an
uncertainty-first pipeline for
four-class brain tumor MRI (glioma, meningioma, pituitary, no tumor) that reads predictive uncertainty
from Monte Carlo (MC) Dropout over $T=20$ passes and turns the resulting entropy into an
explicit rule for deferring uncertain cases to a radiologist. We partitioned
7{,}200 images by perceptual-hash cluster, closing the near-duplicate leakage that inflates accuracy
under naive splitting, and evaluated the pipeline on ViT-B/16 and
ResNet-50 across five seeds along three axes: discrimination, calibration, and selective
prediction. Both discriminate strongly (macro-AUC 0.994;
accuracy 0.962 and 0.964), and no seed separates them ($0$ of $5$ significant, $p<0.05$), so the
result is driven by the uncertainty pipeline, not the network. A single temperature scalar pulls
the deterministic softmax into tight calibration (expected
calibration error 0.016--0.020), and deferring the most uncertain 5\% of cases lifts accuracy on the
rest to $\approx 0.98$ on both (area under the risk--coverage curve 0.010--0.011). MC-Dropout
uncertainty here is thus calibrated, non-collapsing, and directly actionable through a
concrete deferral rule, providing an
architecture-agnostic basis for calibrated, defer-to-human brain tumor MRI triage under internal
validation.

\vspace{0.6em}
\noindent\textbf{Keywords:} brain tumor; magnetic resonance imaging; deep learning; uncertainty
quantification; Monte Carlo dropout; calibration; selective prediction; Vision Transformer;
ResNet; TRIPOD+AI.
\clearpage

\section{Introduction}

Brain tumors are among the most consequential findings in diagnostic neuroimaging. Their subtype on
magnetic resonance imaging (MRI)---glioma, meningioma, pituitary adenoma, or the absence of
tumor---sets the urgency of referral and shapes the initial course of management, which makes
automated classification an attractive way to lighten radiologist workload. Convolutional and
transformer networks now reach accuracies that rival specialist
readers.\citep{he2016,dosovitskiy2021,rokh2025,resnet34_2025} Adoption has lagged those benchmark
numbers anyway. A model that is right 96\% of the time helps at the point of care only if it can also
name the 4\% it is about to get wrong and pass those cases to a human instead of acting on them
silently. What gates deployment, then, is not discrimination but whether the model's own confidence
can be trusted: it has to know when it does not know, and that signal has to be reliable enough to
trigger a defer-to-human decision. We built this study around that requirement, treating calibrated
predictive uncertainty as a primary design objective rather than a diagnostic bolted on after the
fact.

Cheap uncertainty estimates are tempting, and our first one failed outright. An early version of the
project used a deterministic auxiliary head: one extra output, regressed against a confidence target
in the same forward pass as the classifier. It collapsed. Within a few epochs the head was emitting a
nearly constant value that carried almost no information about whether the prediction beneath it was
correct, confident on the errors and the correct cases alike. This is not a quirk of our setup. A
confidence signal trained against the task loss has no reason to track epistemic uncertainty when
minimizing the loss is easier, and the same collapse is documented for deterministic single-network
estimators.\citep{sensoy2018} That failure is what pushed us toward sampling. Among the principled
alternatives---deep ensembles,\citep{lakshminarayanan2017} evidential networks,\citep{sensoy2018} and
Bayesian approximations\citep{kendall2017}---MC Dropout\citep{gal2016} was the one we could add to an
existing backbone at almost no cost: leave dropout on at test time, run the input through $T=20$
times, and take the dispersion of the outputs as the uncertainty signal. It also carries a defensible interpretation as approximate Bayesian
inference, which the auxiliary head never earned. For an input $x$ with dropout masks $\{\mathbf{z}^{(t)}\}_{t=1}^{T}$, the predictive
distribution is the mean of the stochastic softmax outputs,
\begin{equation}
  \bar{\mathbf{p}}(x) \;=\; \frac{1}{T}\sum_{t=1}^{T}
  \operatorname{softmax}\!\big(f_{\theta}(x;\mathbf{z}^{(t)})\big),
  \label{eq:mcmean}
\end{equation}
the reported class is $\hat{y}(x)=\arg\max_{k}\,\bar{p}_k(x)$, and the per-prediction uncertainty is
the predictive entropy of this mean vector,
\begin{equation}
  H\!\big(\bar{\mathbf{p}}(x)\big) \;=\; -\sum_{k=1}^{4}\bar{p}_k(x)\,\log_2 \bar{p}_k(x).
  \label{eq:entropy}
\end{equation}
Taking the base-2 logarithm puts $H$ in bits, so $H\in[0,2]$ for four classes; the base is only a
constant scale and changes neither the ranking nor any deferral decision. And because this entropy
comes from the classifier's own averaged output, it cannot drift free of the task the way a separate
head can.

None of this matters without an honest accuracy baseline underneath it, and here the early version
misled us again. It posted 98.9\% test accuracy under a naive per-file random split. An audit traced
that number straight to the structure of the source data. The meningioma class contains 203 augmented
near-duplicate images, and perceptual hashing across the full dataset found that 1{,}107 of 4{,}784
clusters (23.1\% of clusters) hold more than one near-identical image. Those multi-image clusters
account for 3{,}523 images (48.9\% of the dataset), of which 2{,}416 are redundant near-duplicates
beyond one representative per cluster (33.6\%). A file-level shuffle distributes these duplicates across the
train/test boundary, and the model then scores well by recognizing images it has effectively already
seen. The failure is well known in medical imaging, where slice- and image-level splitting inflates
reported accuracy by enough to flip a study's conclusions.\citep{tampu2022,rumala2023} We avoid it by
grouping on perceptual-hash cluster (Hamming distance $\le 5$) and keeping every near-duplicate of an
image inside a single split; none of the leakage-inflated figures from the earlier split survive into
the analyses below.

So the study we actually report is an uncertainty-first pipeline for four-class brain tumor MRI, one
that turns the MC-Dropout entropy of Eq.~\eqref{eq:entropy} into an explicit threshold for when a case
is deferred to a human reader. Three questions organize the evaluation. First, is the uncertainty
trustworthy? We measure that on a held-out test set with expected calibration error (ECE) and the
Brier score for calibration, and the risk--coverage curve for selective prediction,\citep{geifman2017}
then read a concrete deferral operating point off it. Second, is the discrimination underneath strong
enough to carry the uncertainty layer? We report accuracy, macro-F1, and macro-AUC with bootstrap
confidence intervals across five seeds. Third, does the backbone matter? A per-seed McNemar test
compares ViT-B/16 against ResNet-50; if it cannot tell them apart, the contribution belongs to the
pipeline and not to either network. One planned check, an entropy-based out-of-distribution test, went
unrun because no out-of-distribution image set was available. The deployment we have in mind is
radiologist-in-the-loop: the model triages, defers the least certain cases, and a radiologist makes
the call. It is not built for autonomous triage, and that framing drives the usability discussion
below.

\section{Methods}

\subsection{Data source}

Images are organized under four class folders (glioma, meningioma, pituitary, no tumor), with
1{,}800 images per class (7{,}200 total). Filenames carry \texttt{Te-}/\texttt{Tr-} prefixes from
the original authors' split, plus an augmented-image marker for meningioma (203 files). The data are
the public Kaggle ``Brain Tumor MRI Dataset'' (Version 2) \citep{nickparvar2023}, which the listing
reports as 7{,}200 files, matching our copy exactly, and which the author compiled by merging the
Br35H, SARTAJ, and Figshare brain-tumor sources. We use the images but not the original train/test
split (Section~\ref{sec:analytical}). Two properties of the source are worth stating plainly, as
they bound what the study can claim: the meningioma class includes 203 augmentation-marked files
(rotations/brightness variants of originals), which is what the near-duplicate audit below is built
to contain; and the per-image labels are inherited from the merged sources, whose labeling protocol
and reader qualifications are not published. Acquisition dates and enrollment window are not
retained, as secondary, pre-aggregated public data rarely keeps them.

\subsection{Participants, outcome, and predictors}

Because the data arrive pre-aggregated, the details a clinical reviewer would want, namely which
centers acquired the scans, on what scanners, and at what field strength, are not recoverable, and we
carry that forward into Limitations (Section~\ref{sec:limitations}). Every image in the four folders
was included; nothing was excluded beyond a file-extension check. The label is the four-class tumor
category, one per image, inherited from whoever assembled the source; the labeling protocol and reader
credentials behind those labels are not published, so we cannot independently vouch for them. The
predictor is the whole image, with no hand-crafted features. Preprocessing was deliberately minimal:
resize, ImageNet normalization, and a random horizontal flip during training. We omitted vertical
flips and saturation jitter by design, since axial MRI has no top-bottom symmetry to exploit and the
color in an RGB-cast grayscale scan is not real color to perturb.

\subsection{Sample size and missing data}

The dataset fixes the sample size at 7{,}200, and we ran no prospective power calculation to hit a
target CI width. After the fact, precision was reasonable: on the 1{,}112-image test set, the
bootstrap accuracy CI (percentile, $n=1{,}000$, averaged over seeds) had a half-width near 1.1--1.2
points, and macro-AUC near 0.3--0.4. The McNemar comparison is a different matter. At this test-set
size it is badly underpowered, with achieved power running from 0.000 to 0.285 across seeds
(Section~\ref{sec:comparison}), and the null result there should be read with that in mind. No images
or labels were missing; every file belongs to exactly one class by construction.

\subsection{Analytical methods}
\label{sec:analytical}

\paragraph{Data partitioning.}
A 70/15/15 split was generated by a descending-priority leakage-safety cascade: (1)~patient-ID
parsing from the filename stem; (2)~if patient IDs cannot be parsed for every file---the case
here---perceptual-hash (pHash, Hamming distance $\le 5$) clustering of near-duplicate images
within each class, grouped so near-duplicates do not leak across splits; and (3)~a per-file
stratified fallback only if perceptual hashing is unavailable. Patient-ID parsing matched 0 of
7{,}200 filenames, so step~2 executed. An independent audit confirmed 4{,}784 clusters. Of these,
1{,}107 (23.1\% of \emph{clusters}) hold more than one near-duplicate image. These multi-image
clusters contain 3{,}523 \emph{images} in total (48.9\% of the dataset), of which 2{,}416 are
redundant near-duplicates beyond one representative per cluster (33.6\%; largest cluster, 28
images). The two percentages describe different quantities---images that belong to a multi-image
cluster versus redundant duplicates within them---and are reported together here to avoid confusion.
Observed split sizes are given in
Table~\ref{tab:splits}.

\paragraph{Model class, architecture, and optimization.}
\vit{} uses a torchvision \vit{} backbone with only encoder blocks 10--11, the final encoder
LayerNorm (\texttt{encoder.ln}), and the classification head trainable. \rn{} uses a torchvision \rn{} (ImageNet1K\_V2) with only \texttt{layer4} and the
head trainable, keeping frozen-stage BatchNorm in evaluation mode during training. The shared head
is $\mathrm{LayerNorm}\!\to\!\mathrm{Dropout}(0.3)\!\to\!\mathrm{Linear}(256)\!\to\!\mathrm{GELU}
\!\to\!\mathrm{Dropout}(0.3)\!\to\!\mathrm{Linear}(4)$. Optimization used AdamW\citep{loshchilov2019} (learning rate
$1\times10^{-4}$, weight decay 0.01) with ReduceLROnPlateau, label smoothing 0.1, early stopping
(patience 5) on validation loss, and up to 30 epochs, over five seeds.

\paragraph{Performance measures.}
Discrimination used accuracy, macro-F1, and macro-AUC (one-vs-rest). Calibration used the expected
calibration error, computed by partitioning the $N$ test predictions into $M=15$ equal-width
confidence bins $\{B_m\}$,
\begin{equation}
  \ece \;=\; \sum_{m=1}^{M} \frac{|B_m|}{N}\,
  \big|\,\mathrm{acc}(B_m) - \mathrm{conf}(B_m)\,\big|,
  \label{eq:ece}
\end{equation}
where $\mathrm{acc}(B_m)$ and $\mathrm{conf}(B_m)$ are the mean accuracy and mean predicted
confidence within bin $B_m$; and the multiclass Brier score,
\begin{equation}
  \mathrm{Brier} \;=\; \frac{1}{N}\sum_{i=1}^{N}\sum_{k=1}^{4}
  \big(\bar{p}_{ik} - y_{ik}\big)^2,
  \label{eq:brier}
\end{equation}
with $y_{ik}$ the one-hot ground truth. Selective prediction used the risk--coverage curve:
predictions are ranked by the entropy of Eq.~\eqref{eq:entropy}, and for a coverage
$\tau\in(0,1]$ the retained set $S_\tau$ is the fraction $\tau$ of lowest-entropy predictions,
with risk
\begin{equation}
  \mathrm{risk}(\tau) \;=\; \frac{1}{|S_\tau|}\sum_{i\in S_\tau}
  \mathbf{1}\!\left[\hat{y}_i \neq y_i\right],
  \qquad
  \mathrm{AURC} \;=\; \int_{0}^{1}\mathrm{risk}(\tau)\,d\tau,
  \label{eq:aurc}
\end{equation}
approximated by the trapezoidal rule over the sorted predictions; accuracy at fixed coverage
(80/90/95\%) is reported alongside. Architecture comparison used the continuity-corrected McNemar
test on the discordant pairs $(b,c)$---examples where exactly one architecture is correct---
\begin{equation}
  \chi^2 \;=\; \frac{\big(|b-c|-1\big)^2}{b+c},
  \label{eq:mcnemar}
\end{equation}
evaluated per seed. All 95\% confidence intervals use the percentile bootstrap with $n=1{,}000$
resamples: for a metric $\phi$ and resamples $\phi^{*}_{(1)}\le\dots\le\phi^{*}_{(n)}$, the
interval is $\big[\phi^{*}_{(\lfloor 0.025 n\rfloor)},\,\phi^{*}_{(\lceil 0.975 n\rceil)}\big]$.

\paragraph{Model output.}
The output is the four-way MC-Dropout mean of Eq.~\eqref{eq:mcmean}; the reported class is its
argmax, with the entropy of Eq.~\eqref{eq:entropy} as the uncertainty score. No fixed decision
threshold beyond argmax is used; coverage-based deferral thresholds are reported at fixed coverage
levels, not as a single deployment recommendation.

\subsection{Class imbalance, fairness, and ethics}

The dataset is balanced by construction (1{,}800 images/class); no imbalance correction was needed.
No demographic, site, or scanner metadata is available, so no fairness or subgroup evaluation was
performed; this is carried into Limitations. This study used a publicly available, de-identified
imaging dataset and did not involve any new human-subjects data collection; separate institutional
review board approval was therefore not required.

\section{Results}

\subsection{A leakage-safe partition of 7,200 brain MRI images}

\begin{figure}[H]
  \centering
  \includegraphics[width=0.82\textwidth,height=0.80\textheight,keepaspectratio]{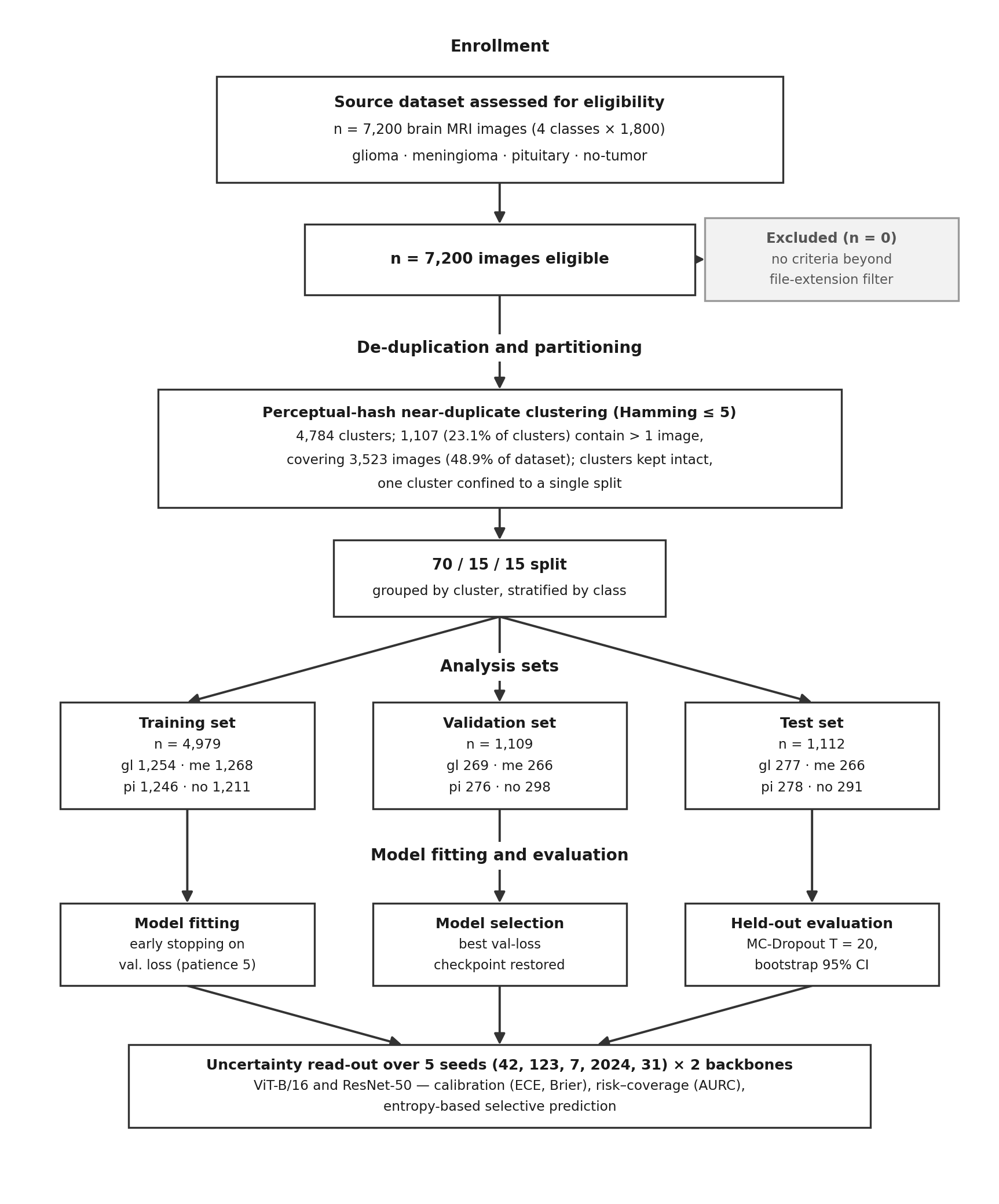}
  \caption{Image flow and dataset partition. The 7{,}200-image source dataset (four classes
  $\times$ 1{,}800) was grouped by perceptual-hash near-duplicate cluster (Hamming distance
  $\le 5$) so that all near-duplicates of an image remain within a single split, then partitioned
  70/15/15. Models were fitted with early stopping on validation loss, selected by best
  validation-loss checkpoint, and evaluated once on the held-out test set with MC-Dropout
  ($T=20$) across five seeds and two backbones.}
  \label{fig:flow}
\end{figure}

We partitioned the 7{,}200 images so that no near-duplicate could straddle the train/test boundary.
Every image entered the analysis; nothing was dropped beyond a file-extension check. Grouping on
perceptual-hash cluster and splitting 70/15/15 sent 4{,}979 images to training, 1{,}109 to validation,
and 1{,}112 to a held-out test set that never touched a training decision (Figure~\ref{fig:flow},
Table~\ref{tab:splits}). Class balance held in every split (Table~\ref{tab:splits}). All numbers below
come from that test set.

\begin{table}[H]
  \centering
  \caption{Split sizes by class from the perceptual-hash cluster-grouped 70/15/15 partition.
  Counts are deterministic and taken from the generated split manifest.}
  \label{tab:splits}
  \begin{tabular}{lccccc}
    \toprule
    Split & glioma & meningioma & pituitary & no tumor & \textbf{Total} \\
    \midrule
    train      & 1{,}254 & 1{,}268 & 1{,}246 & 1{,}211 & \textbf{4{,}979} \\
    validation & 269     & 266     & 276     & 298     & \textbf{1{,}109} \\
    test       & 277     & 266     & 278     & 291     & \textbf{1{,}112} \\
    \bottomrule
  \end{tabular}
\end{table}

\subsection{Both backbones train stably and discriminate strongly}

Both backbones trained across five seeds, with only the upper layers and a shared classification head
left unfrozen (Methods). Training was stable and stopped early. Nine of the ten runs halted well
short of the 30-epoch budget under patience-5 early stopping on validation loss; only ResNet-50 seed
42 ran the full 29 epochs, and the rest halted between epochs 10 and 18 (Table~\ref{tab:earlystop}). We
restored each model's best validation-loss checkpoint before touching the test set.

\begin{table}[H]
  \centering
  \caption{Early-stopping behavior by model and seed. Best epoch is the lowest-validation-loss
  epoch, whose weights were restored for test evaluation.}
  \label{tab:earlystop}
  \begin{tabular}{llcccc}
    \toprule
    Model & Seed & Stopped at & Best epoch & Best val-loss & Best val-acc \\
    \midrule
    \vit{} & 42   & 10 & 5  & 0.4054 & 0.9757 \\
    \vit{} & 123  & 13 & 8  & 0.4019 & 0.9739 \\
    \vit{} & 7    & 12 & 7  & 0.4114 & 0.9702 \\
    \vit{} & 2024 & 18 & 13 & 0.4038 & 0.9784 \\
    \vit{} & 31   & 10 & 5  & 0.4152 & 0.9693 \\
    \rn{}  & 42   & 29 & 24 & 0.3996 & 0.9748 \\
    \rn{}  & 123  & 13 & 8  & 0.4061 & 0.9720 \\
    \rn{}  & 7    & 14 & 9  & 0.4081 & 0.9748 \\
    \rn{}  & 2024 & 12 & 7  & 0.4024 & 0.9784 \\
    \rn{}  & 31   & 12 & 7  & 0.4227 & 0.9693 \\
    \bottomrule
  \end{tabular}
\end{table}

\subsection{Discrimination is strong and essentially equal across architectures}

\begin{figure}[H]
  \centering
  \includegraphics[width=0.95\textwidth]{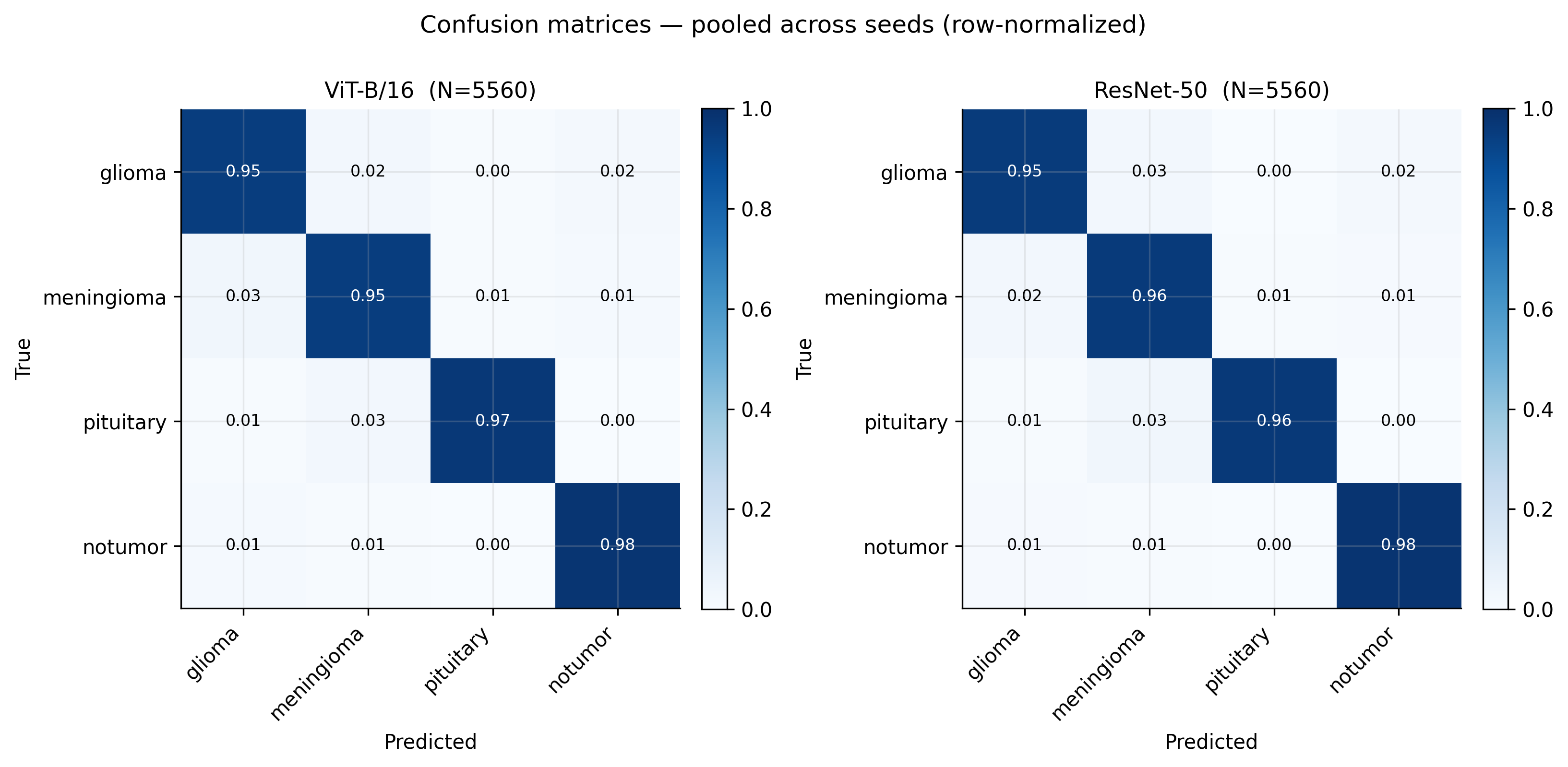}
  \caption{Confusion matrices on the held-out test split (row-normalized, pooled across seeds).
  Off-diagonal errors are concentrated at the glioma--meningioma boundary; the no-tumor class is
  separated most cleanly.}
  \label{fig:cm}
\end{figure}

\begin{figure}[H]
  \centering
  \includegraphics[width=0.95\textwidth]{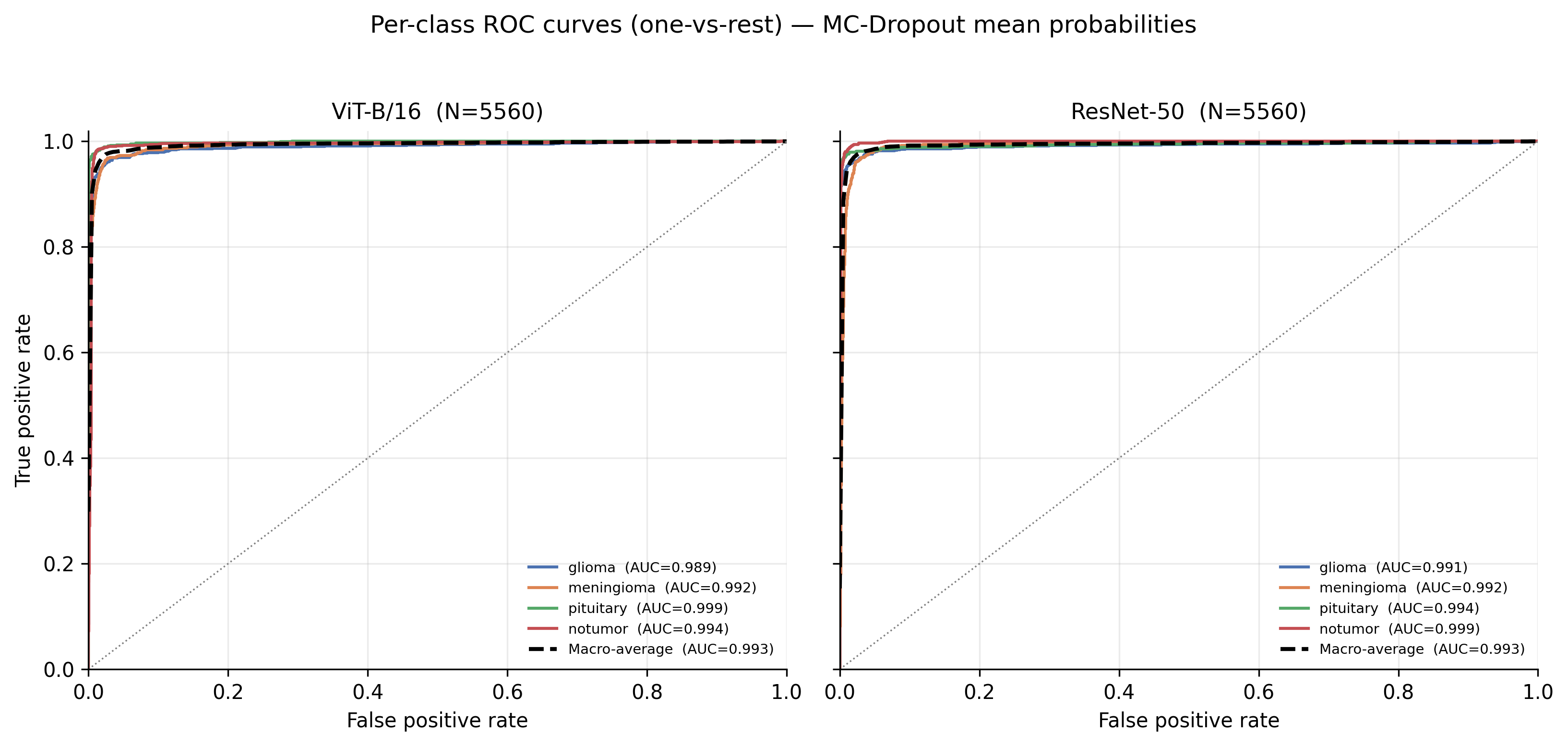}
  \caption{Per-class one-vs-rest ROC curves with per-class AUC. All four classes exceed
  AUC $\approx 0.99$, and the two architectures are visually indistinguishable. The macro-AUC shown
  in the figure legend is computed by pooling the predictions of all five seeds before averaging and
  rounds to 0.993; this differs slightly from the per-seed mean macro-AUC in Table~\ref{tab:perf}
  (0.994), which pools and averages in the opposite order. Both summarize the same underlying
  discrimination; the Table~\ref{tab:perf} per-seed means are the values used in the text.}
  \label{fig:roc}
\end{figure}

On the held-out test set, both backbones pulled the four tumor classes apart cleanly. Averaged over
the five seeds, ViT-B/16 and ResNet-50 each hit a macro-AUC of 0.994
(Figures~\ref{fig:cm}--\ref{fig:roc}, Table~\ref{tab:perf}), at mean accuracies of 0.962 and 0.964.
Where they erred was the glioma--meningioma boundary; the no-tumor class was the easiest to
isolate (Figure~\ref{fig:cm}). Calibration under MC-Dropout averaging was middling, with an ECE of
0.070--0.074 (Figure~\ref{fig:cal}). Selective prediction closed that gap: withhold the most uncertain
5\% of cases and accuracy on the rest climbs to 0.980 (Figure~\ref{fig:rc}).

\begin{table}[H]
  \centering
  \caption{Across-seed mean test performance (five seeds). CIs are the mean of per-seed bootstrap
  95\% CI bounds ($n=1{,}000$). OOD-AUROC was not run.}
  \label{tab:perf}
  \setlength{\tabcolsep}{4pt}
  \small
  \begin{tabular}{lccccccc}
    \toprule
    Model & Accuracy (95\% CI) & Macro-F1 & Macro-AUC (95\% CI) & \ece & Brier & AURC & Acc@95\% \\
    \midrule
    \vit{} & 0.9615 (0.950--0.973) & 0.9616 & 0.9938 (0.990--0.997) & 0.0699 & 0.0677 & 0.0102 & 0.9797 \\
    \rn{}  & 0.9642 (0.953--0.975) & 0.9638 & 0.9939 (0.990--0.997) & 0.0738 & 0.0659 & 0.0105 & 0.9801 \\
    \bottomrule
  \end{tabular}
\end{table}

\begin{table}[H]
  \centering
  \caption{Per-seed values.}
  \label{tab:perseed}
  \setlength{\tabcolsep}{5pt}
  \begin{tabular}{llcccccc}
    \toprule
    Model & Seed & Accuracy & Macro-F1 & Macro-AUC & \ece & Brier & AURC \\
    \midrule
    \vit{} & 42   & 0.9631 & 0.9627 & 0.9948 & 0.0812 & 0.0689 & 0.0079 \\
    \vit{} & 123  & 0.9640 & 0.9639 & 0.9953 & 0.0721 & 0.0663 & 0.0058 \\
    \vit{} & 7    & 0.9613 & 0.9610 & 0.9933 & 0.0673 & 0.0640 & 0.0093 \\
    \vit{} & 2024 & 0.9622 & 0.9621 & 0.9919 & 0.0647 & 0.0661 & 0.0174 \\
    \vit{} & 31   & 0.9568 & 0.9580 & 0.9936 & 0.0643 & 0.0733 & 0.0108 \\
    \rn{}  & 42   & 0.9649 & 0.9648 & 0.9909 & 0.0680 & 0.0652 & 0.0204 \\
    \rn{}  & 123  & 0.9604 & 0.9600 & 0.9947 & 0.0757 & 0.0697 & 0.0089 \\
    \rn{}  & 7    & 0.9649 & 0.9647 & 0.9971 & 0.0731 & 0.0594 & 0.0027 \\
    \rn{}  & 2024 & 0.9667 & 0.9665 & 0.9954 & 0.0750 & 0.0630 & 0.0072 \\
    \rn{}  & 31   & 0.9640 & 0.9628 & 0.9916 & 0.0774 & 0.0721 & 0.0134 \\
    \bottomrule
  \end{tabular}
\end{table}

\subsection{A single temperature restores calibration without changing predictions}

\begin{figure}[H]
  \centering
  \includegraphics[width=0.92\textwidth]{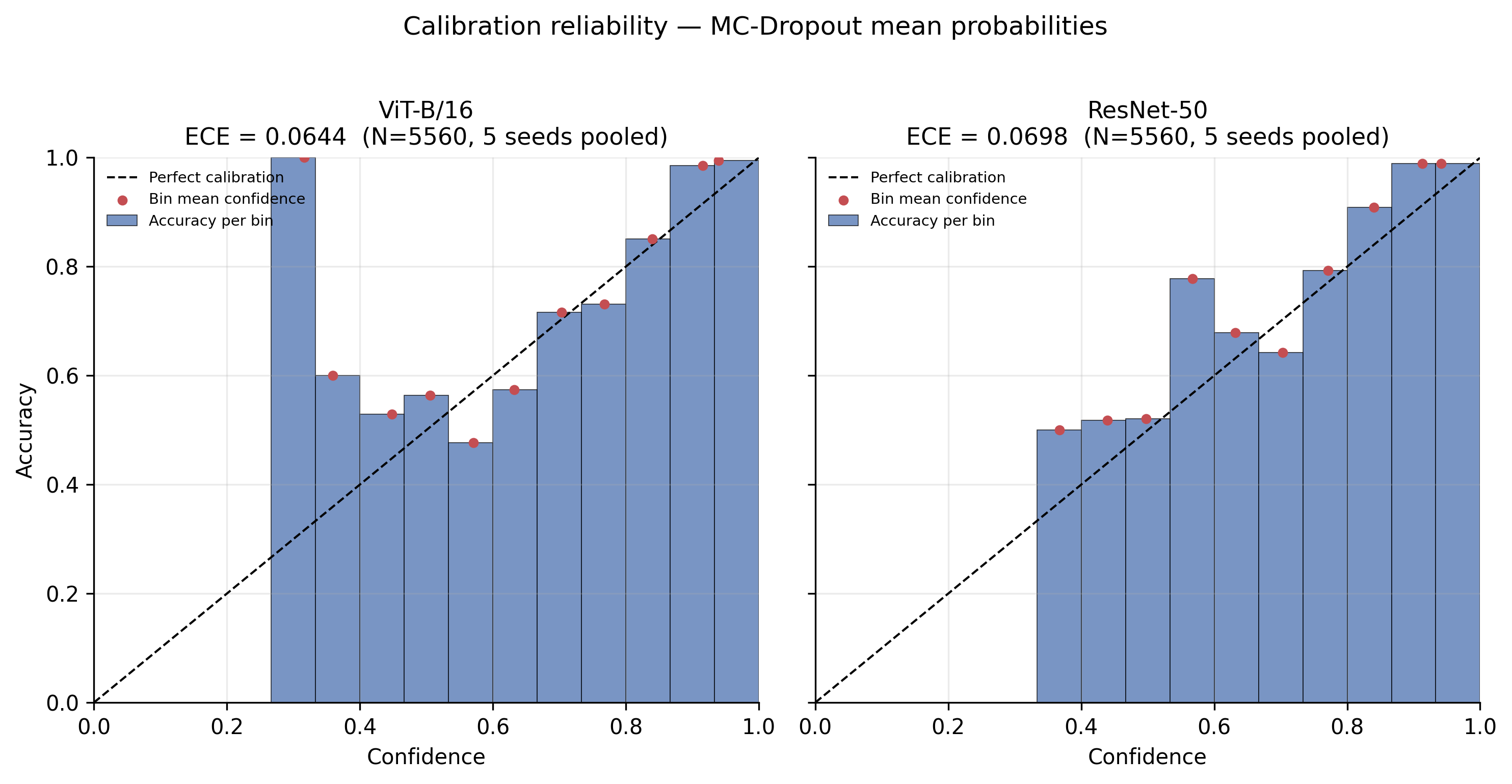}
  \caption{Reliability diagrams (MC-Dropout mean probability). Observed accuracy versus predicted
  confidence over 15 equal-width bins; bars above the diagonal indicate under-confidence (accuracy
  exceeds stated confidence), as seen here. The ECE
  annotated in each panel (0.0644 for ViT-B/16, 0.0698 for ResNet-50) is computed by pooling the
  predictions of all five seeds before binning. This differs slightly from the ECE in
  Table~\ref{tab:perf} (0.0699 and 0.0738), which is the mean of the five per-seed ECE values;
  pooling and per-seed averaging are distinct calculations, and both are reported. The
  Table~\ref{tab:perf} per-seed means are the values used throughout the text.}
  \label{fig:cal}
\end{figure}

\begin{figure}[H]
  \centering
  \includegraphics[width=0.6\textwidth]{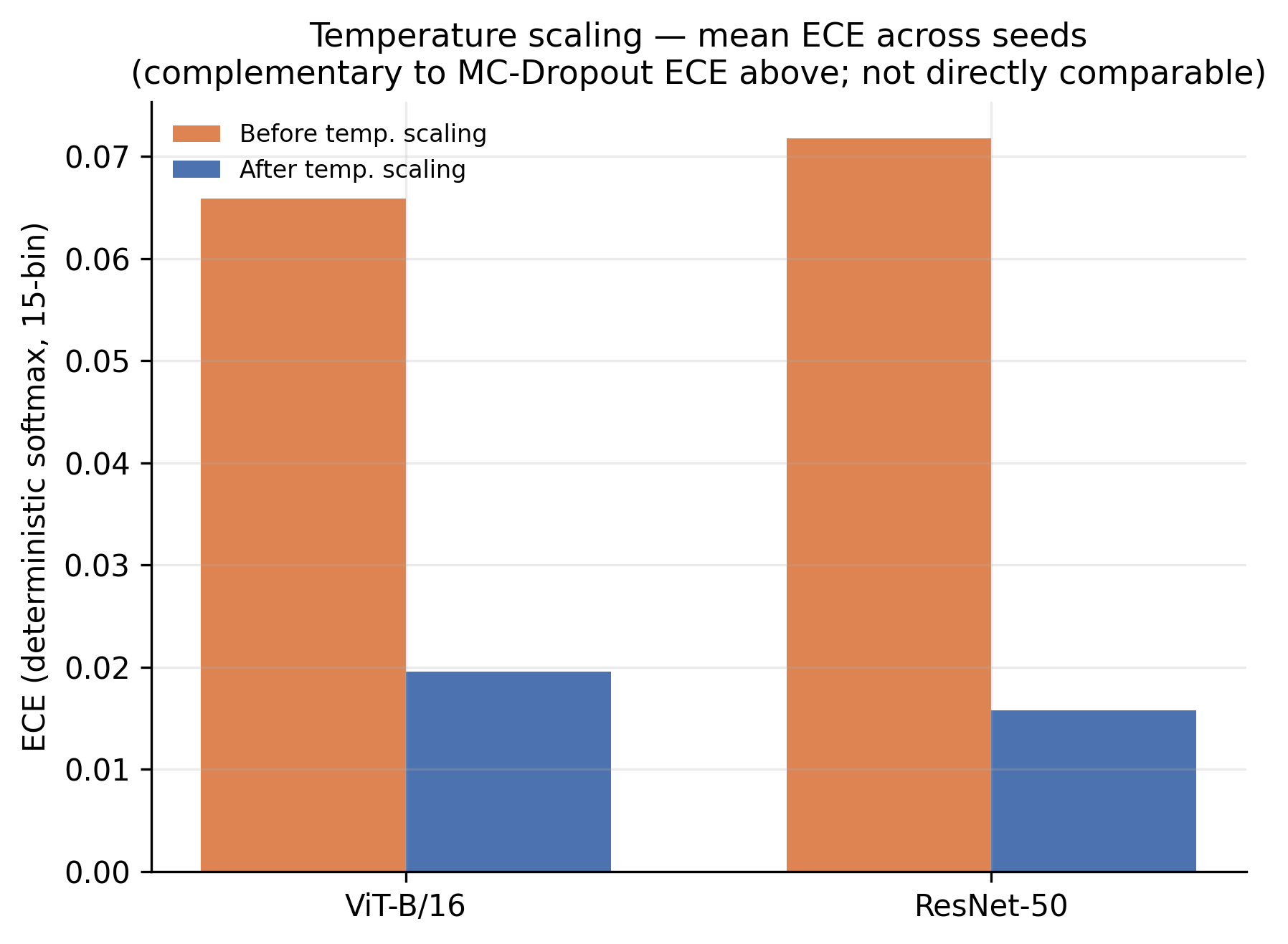}
  \caption{Effect of post-hoc temperature scaling on the deterministic softmax, fitted on the
  validation split \citep{guo2017}. Fitted temperature averaged 0.62; the deterministic-softmax ECE
  fell from $\approx$0.066--0.072 to $\approx$0.016--0.020 without any change to accuracy. These
  single-pass ECE values are distinct from the MC-Dropout ECE in Table~\ref{tab:perf} (see \S3.4).}
  \label{fig:temp}
\end{figure}

The reliability diagrams tell the same story as the ECE: both models sit just above the diagonal---their
accuracy modestly exceeds their stated confidence, meaning they are mildly \emph{under}-confident
(Figure~\ref{fig:cal}). This is readily corrected. Fitting one temperature scalar on the validation
split's deterministic (dropout-off) logits\citep{guo2017} cut the \emph{deterministic-softmax} test ECE
from about 0.066 to 0.020 for ViT-B/16 and from 0.072 to 0.016 for ResNet-50, at an average temperature
of 0.62---a value below 1, which sharpens the softmax to raise its confidence, exactly the direction
under-confidence calls for (Figure~\ref{fig:temp}). A note on which ECE this is: these values come from
the single-pass (dropout-off) softmax, so they are distinct from, and a little lower than, the
MC-Dropout ECE in Table~\ref{tab:perf} (0.0699 and 0.0738), which is computed on the $T=20$ MC-averaged
distribution. Temperature scaling only rescales the logits, so it leaves the argmax---and the
accuracy---untouched. The two measurements are complementary, not contradictory: they live on different
predictive distributions, and both say the same thing, that the raw output is mildly under-confident
and that either MC-Dropout averaging or a single temperature restores calibration.

\subsection{Entropy-ranked deferral yields a usable operating point}

\begin{figure}[H]
  \centering
  \includegraphics[width=0.62\textwidth]{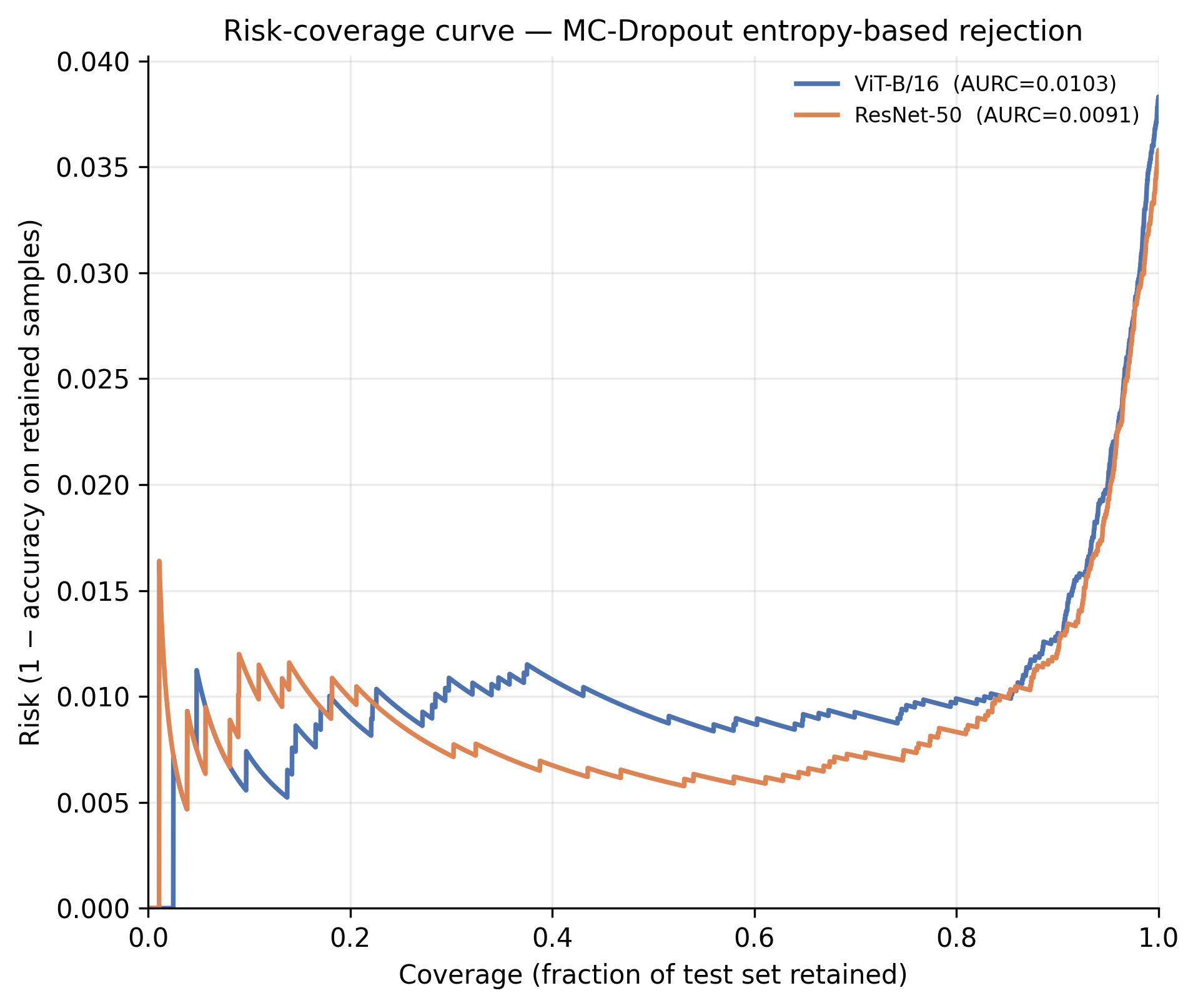}
  \caption{Risk--coverage curves. Test-set error as a function of coverage under entropy-ranked
  deferral, for both backbones and across seeds. Lower is better. The authoritative area under the
  risk--coverage curve (AURC), computed per seed and averaged, is reported in
  Table~\ref{tab:perf} (0.0102 for ViT-B/16 and 0.0105 for ResNet-50) and matches the per-seed
  values in Table~\ref{tab:perseed}. The AURC values annotated in the figure legend (0.0103 and
  0.0091) derive from a separate inference pass and differ slightly because the MC-Dropout read-out
  is not re-seeded per evaluation call (Appendix~\ref{app:repro}); the discrepancy is well within
  the run-to-run variation described there and does not affect any reported conclusion. The
  Table~\ref{tab:perf} values should be treated as authoritative.}
  \label{fig:rc}
\end{figure}

The entropy of Eq.~\eqref{eq:entropy} works well as a deferral signal. Rank the test cases by entropy,
withhold the most uncertain slice, and the risk--coverage curve comes out favourable for both
backbones (Figure~\ref{fig:rc}): at 95\% coverage, accuracy on what remains rises to 0.980, up from
0.962 (ViT-B/16) and 0.964 (ResNet-50) at full coverage. The area under that curve is small for both
(AURC $\approx 0.010$--0.011). Stated plainly, deferring one case in twenty recovers close to two points
of accuracy on the rest---a concrete operating point for a defer-to-human workflow.

\subsection{No backbone outperforms the other on any seed}
\label{sec:comparison}

\begin{figure}[H]
  \centering
  \includegraphics[width=0.75\textwidth]{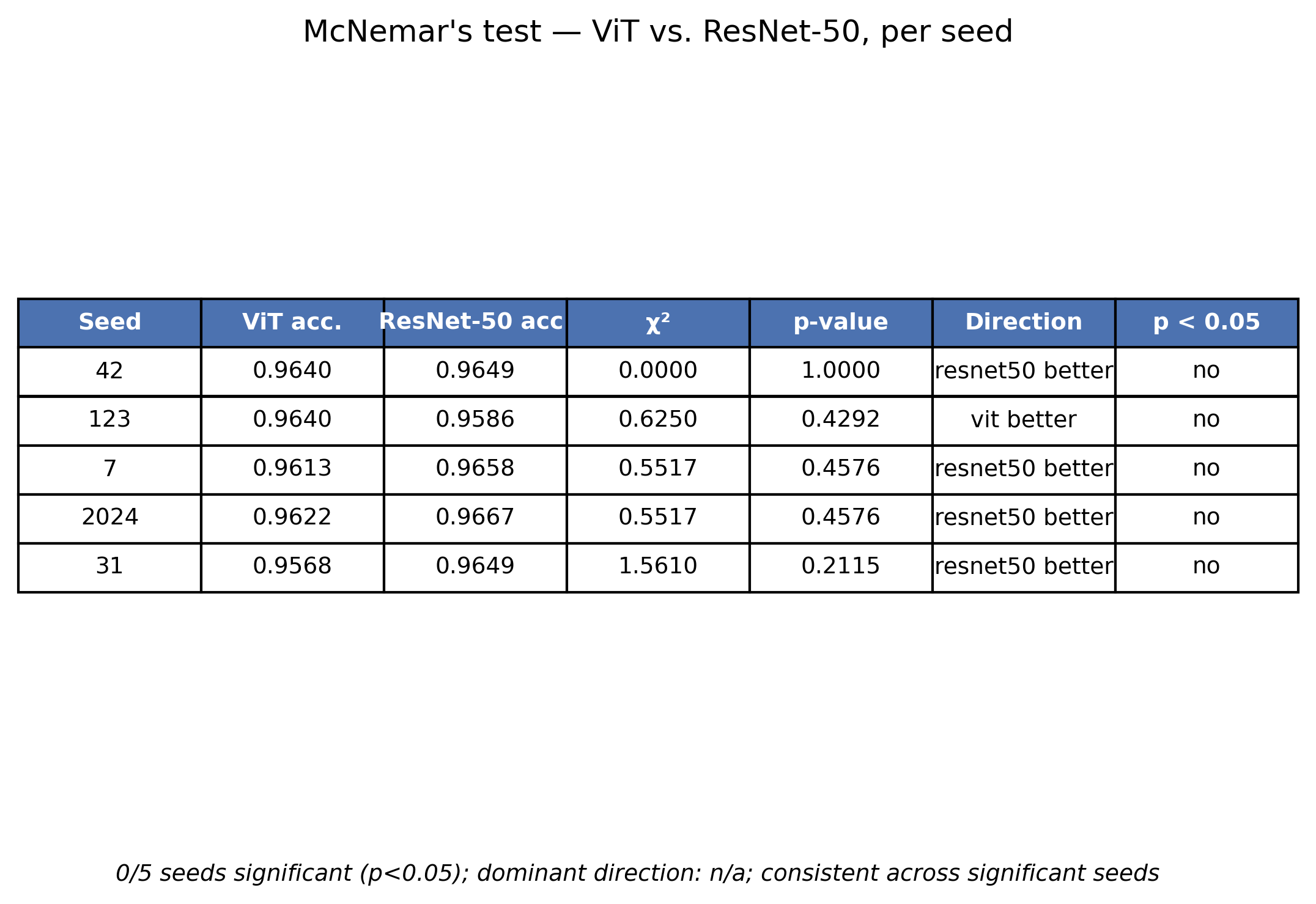}
  \caption{Architecture comparison summary. Per-seed McNemar results rendered as a panel. No seed
  reaches significance; the effect direction is inconsistent across seeds.}
  \label{fig:mcnemar}
\end{figure}

We compared the two backbones head-to-head on the same test set with a per-seed McNemar test
(Eq.~\eqref{eq:mcnemar}). Not one seed reached significance (0 of 5 at $p<0.05$;
Table~\ref{tab:mcnemar}, Figure~\ref{fig:mcnemar}). ResNet-50 was nominally ahead on four seeds and
ViT-B/16 on one, but the direction flipped from seed to seed and nothing came close ($p$-values
0.211--1.000; $\chi^2$ 0.00--1.56). The test is also underpowered at this sample size (achieved power
0.000--0.285; Section~\ref{sec:analytical}), so what we have is an absence of detectable difference,
not a proof that the two are equal. Either way the reading is the same: the two architectures behave
alike on discrimination, calibration, and deferral, which puts the source of the result in the
uncertainty pipeline rather than the network carrying it.

\begin{table}[H]
  \centering
  \caption{Per-seed McNemar comparison of \vit{} versus \rn{} on the identical test set
  (continuity-corrected).}
  \label{tab:mcnemar}
  \begin{tabular}{lccccl}
    \toprule
    Seed & \vit{} acc & \rn{} acc & $\chi^2$ & $p$-value & Direction \\
    \midrule
    42   & 0.9640 & 0.9649 & 0.0000 & 1.0000 & \rn{} better \\
    123  & 0.9640 & 0.9586 & 0.6250 & 0.4292 & \vit{} better \\
    7    & 0.9613 & 0.9658 & 0.5517 & 0.4576 & \rn{} better \\
    2024 & 0.9622 & 0.9667 & 0.5517 & 0.4576 & \rn{} better \\
    31   & 0.9568 & 0.9649 & 1.5610 & 0.2115 & \rn{} better \\
    \bottomrule
  \end{tabular}
\end{table}

\subsection{Attention concentrates on the lesion, and disperses when the model errs}

\begin{figure}[p]
  \centering
  \includegraphics[width=\textwidth,height=0.86\textheight,keepaspectratio]{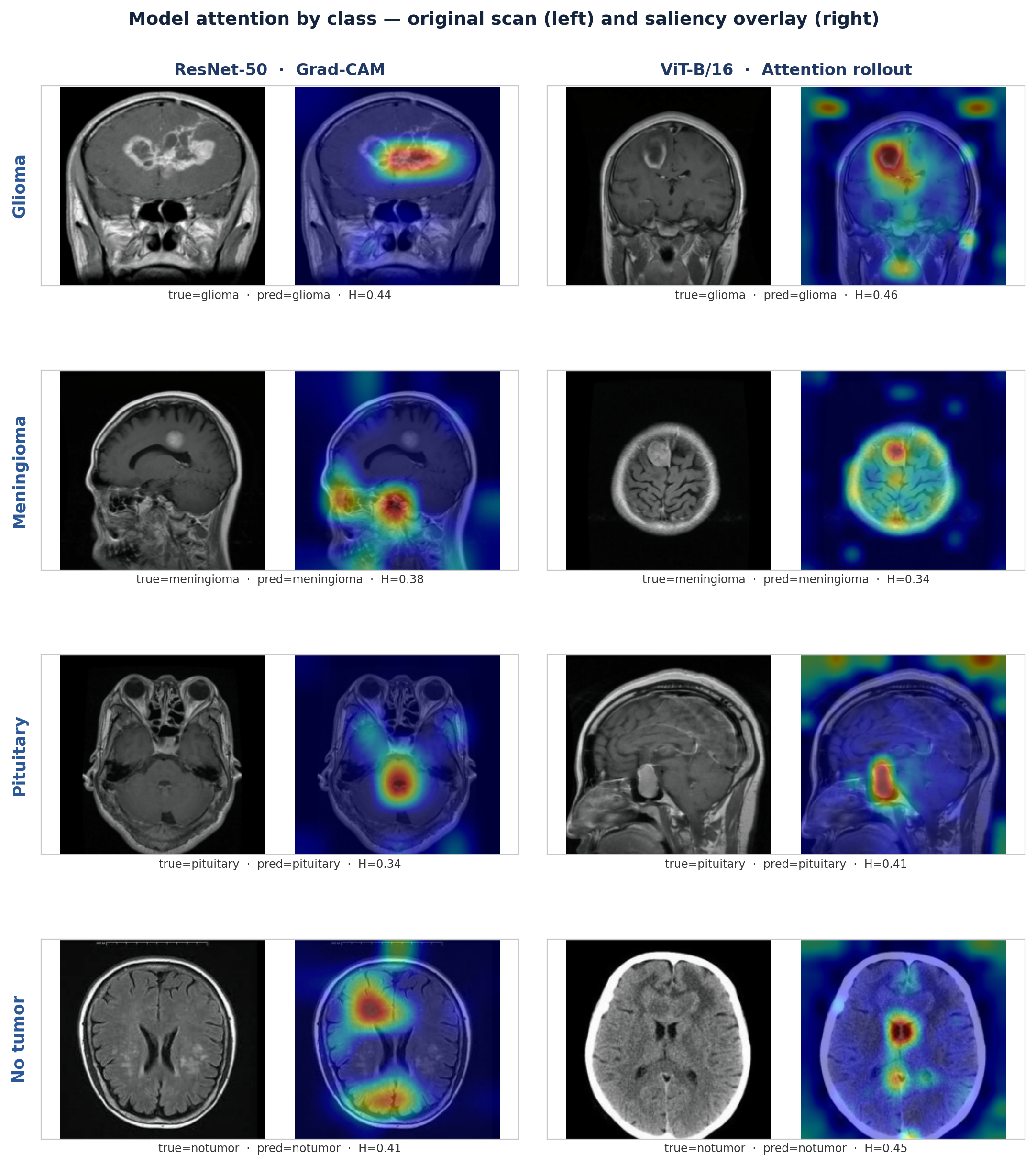}
  \caption{Model attention by class. Each cell pairs the input scan (left) with a saliency overlay
  (right); rows are the four classes, the left column is \rn{} (Grad-CAM) and the right column is
  \vit{} (attention rollout). All eight examples shown are confident, correctly classified test
  cases (predictive entropy $H$ noted beneath each). Warmer colors mark higher attribution. For the
  tumor-positive classes the heat concentrates on the lesion. These maps are qualitative
  interpretability aids and were not used in any quantitative claim.}
  \label{fig:gradcam}
\end{figure}

To see where each model looked, we generated Grad-CAM maps\citep{selvaraju2017} for ResNet-50 and
attention-rollout maps\citep{abnar2020} for ViT-B/16 across all four classes (Figure~\ref{fig:gradcam}).
On confident, correctly classified tumor-positive cases, the attribution sat on the lesion
(Figure~\ref{fig:gradcam}). The pattern held across the wider pool, and the misses were the telling
part: when a model got a case wrong, its attention tended to scatter across the image or slide off the
lesion entirely---the same cases that carry the high predictive entropy the deferral rule uses to set
them aside.

\section{Discussion}

\subsection{Calibrated, actionable uncertainty from an architecture-agnostic pipeline}

On the internal test set, both networks classified the four tumor types well, macro-AUC near 0.994 and
mean accuracy 96.2--96.4\% across five seeds, and no test we ran separated them. But the accuracy is
not the finding; the uncertainty is. Raw MC-Dropout probabilities were only moderately calibrated (ECE
0.070--0.074, Brier 0.066--0.068), yet a single temperature scaling pulled the deterministic softmax
into tight calibration (ECE 0.016--0.020) without moving a single prediction, and entropy-ranked
deferral converted that uncertainty into a usable operating point, 0.980 accuracy on the retained
cases once the most uncertain 5\% were set aside. For the choice between a convolutional and a
transformer backbone, the implication is deflationary: the architecture is not the operative variable.
At this scale and on these data, performance does not reside in the network. MC Dropout supplies
calibrated, non-collapsing, actionable uncertainty on either backbone, and in a triage setting that is
the property on which a clinician relies.

One caveat qualifies everything above. The source dataset ships with no demographic or site metadata,
so we cannot break performance down by subgroup, and every number we report is an average over a
patient mix we never observed.

\subsection{Relation to prior work}

We place this work against the uncertainty and calibration literature, not the brain-tumor accuracy
leaderboard, because that is where its contribution sits. The MC-Dropout formulation is Gal and
Ghahramani's;\citep{gal2016} what we add is the task-specific demonstration that its entropy is usable
for deferral (AURC 0.010--0.011, accuracy $\approx 0.98$ at 95\% coverage) in a setting where a cheaper
deterministic confidence head collapsed, a failure the evidential-learning
literature\citep{sensoy2018} pins on the overconfidence of loss-trained softmax outputs. On
calibration, a single temperature largely corrects the miscalibration\citep{guo2017}; here the raw
softmax is mildly \emph{under}-confident (fitted $T<1$), the reverse of the overconfidence Guo et al.\
report---consistent with our use of label smoothing (0.1), which caps the true-class target and pushes
confidence below accuracy---but temperature scaling corrects miscalibration in either direction, taking
the deterministic-softmax ECE from 0.066--0.072 down to 0.016--0.020. The selective-prediction framing, trading coverage
for accuracy by deferring on a confidence score, is Geifman and El-Yaniv's,\citep{geifman2017} and deep
ensembles remain the main alternative route to the calibrated uncertainty we get from
sampling.\citep{lakshminarayanan2017}

The contrast with prior brain-tumor MRI classifiers is the more instructive comparison. Recent
transfer-learning studies on the same public dataset family report headline accuracies of 98--99\%,
higher than ours,\citep{rokh2025,resnet34_2025} but they get there under image-level rather than
patient- or duplicate-grouped splits, and they carry no calibration or selective-prediction analysis
at all, reporting a bare softmax confidence at most. Since slice- and image-level leakage is known to
inflate accuracy by margins that can reverse a study's conclusions,\citep{tampu2022,rumala2023} those
figures do not line up against a leakage-controlled evaluation, and they leave the one question a
clinician actually faces---when to trust the prediction---untouched. Set against that, the advance
here is not a bigger accuracy number but an explicit, calibrated risk--coverage operating point earned
on a leakage-controlled split.

\subsection{Limitations}
\label{sec:limitations}

A few limitations bound what we can claim. Everything comes from one aggregated public source with no
independent scanner or site variation, so the numbers describe in-distribution behaviour and nothing
more; there is no second imaging source anywhere in the study, and external validation on an
independent cohort---the reason we keep calling this internal validation---is still to come. The data
are retrospective, assembled after the fact from a public release, and we cannot audit where the
labels came from. With no demographic or site metadata in the source, fairness across subgroups is
simply untestable here, and that gap is not a footnote: it touches every value we report. We also ran
no reader study and no workflow evaluation, so the claim stops at a candidate decision-support
component with calibrated internal-validation performance.

One limitation bears on the central claim more directly than the rest. Perceptual-hash grouping stops
\emph{near-duplicate images} from crossing the split, but the source hands us no patient identifiers,
so two genuinely different slices from one patient can still land on opposite sides of the partition.
Leakage like that is usually framed as a threat to accuracy; here it threatens the very thing we most
want to trust. If train and test share patients, the held-out entropy and ECE could look better
calibrated than they would on truly unseen patients, and the 95\%-coverage deferral point could be
optimistic. The per-seed McNemar comparison is underpowered on top of that (achieved power
0.000--0.285), so its null is an absence of detectable difference, not a demonstration of
equivalence.

\subsection{Usability and future directions}

With no out-of-distribution image set available, we report no OOD detection metric. The risk--coverage
results still hand us a working deferral rule: at 95\% coverage, retained-case accuracy reached about
0.980 on both backbones, against 0.962--0.964 at full coverage. We offer it as a candidate operating
point, not a validated decision rule, since we tested it only against in-distribution predictive
entropy and never against genuinely atypical or degraded input. The user we have in mind is a
radiologist in the loop: the pipeline surfaces a prediction with a calibrated uncertainty and defers
the least certain cases for human review. It is not meant to triage on its own. The most important
next step is external validation on an independent cohort. A low-cost first step, feasible within the
present data, is to split the merged dataset by its three constituent sub-sources and report
performance per source. After that, a prospective reader study would show whether the calibrated
uncertainty actually shifts clinician decisions, and a dataset that carries demographic or site
metadata would open up the fairness and subgroup analysis this source shuts out.

\section{Conclusion}

MC-Dropout uncertainty for four-class brain tumor MRI is calibrated, non-collapsing, and actionable.
We built an entropy-thresholded selective-prediction pipeline around that fact, validated it
internally on two backbones---ViT-B/16 and ResNet-50---under a leakage-controlled perceptual-hash
partition, and found the two nearly indistinguishable across five seeds: macro-AUC $\approx 0.994$, ECE
in the 0.07--0.08 band, no backbone reliably ahead on any seed. Entropy-ranked deferral holds accuracy
at $\approx 0.98$ once the most uncertain 5\% of cases are withheld. That the uncertainty pipeline, and
not the choice of network, carries the result is, for triage, the central point. What these data cannot
yet establish is whether any of it survives outside a single public dataset. External validation on an
independent cohort is the next step, and the one the field will---rightly---insist on.

\section*{Open Science and Transparency}
\addcontentsline{toc}{section}{Open Science and Transparency}
\textbf{Funding.} This research received no external funding.\;
\textbf{Conflicts of interest.} The author declares no competing interests.\;
\textbf{Protocol/registration.} This study was not pre-registered.\;
\textbf{Data and code availability.} The analysis code, split-generation logic, training pipeline,
and figure scripts are available from the author on reasonable request; the ten trained checkpoints
(2 architectures $\times$ 5 seeds) are likewise available from the author on reasonable request. The
dataset is the public Kaggle ``Brain Tumor MRI Dataset'' (Version 2, 7{,}200 files)
\citep{nickparvar2023}, distributed under the Creative Commons Attribution 4.0 International
(CC BY 4.0) license (\url{https://creativecommons.org/licenses/by/4.0/}).\;
\textbf{Patient and public involvement.} None (retrospective secondary-data analysis).\;
\textbf{Use of AI tools.} During the preparation of this manuscript, the author used Claude to
improve the language, clarity, and grammatical flow of the text. After using this tool, the author
reviewed and edited the content as needed and takes full responsibility for the content of the
publication.


\appendix
\section{Metric Definitions}

Macro-F1 is the unweighted mean of the four one-vs-rest F1 scores; for class $k$ with precision
$P_k$ and recall $R_k$,
\begin{equation}
  \mathrm{F1}_k = \frac{2P_kR_k}{P_k+R_k},
  \qquad
  \text{macro-F1} = \frac{1}{4}\sum_{k=1}^{4}\mathrm{F1}_k .
  \label{eq:f1}
\end{equation}
Macro-AUC is the unweighted mean of the four one-vs-rest areas under the ROC curve. Accuracy at
coverage $\tau$ is $1-\mathrm{risk}(\tau)$ from Eq.~\eqref{eq:aurc}. Temperature scaling fits a
single scalar $T>0$ by minimizing validation negative log-likelihood,
\begin{equation}
  T^{\star} = \arg\min_{T>0}\; -\sum_{i}\log
  \Big[\operatorname{softmax}\!\big(\mathbf{z}_i / T\big)\Big]_{y_i},
  \label{eq:temp}
\end{equation}
where $\mathbf{z}_i$ are the deterministic (single-pass) logits; $T^{\star}$ is then applied to the
test logits.

\section{Reproducibility}
\label{app:repro}

Five seeds (42, 123, 7, 2024, 31) fix all random-number generators (\texttt{random}, \texttt{numpy},
\texttt{torch} CPU/CUDA), set \texttt{cudnn.deterministic = True} and \texttt{cudnn.benchmark =
False}, and \texttt{PYTHONHASHSEED}, for each run independently. Training used a 30-epoch budget with
early stopping (patience 5), MC-Dropout $T=20$ at evaluation, bootstrap 95\% CIs ($n=1{,}000$), and
the per-seed McNemar test of Eq.~\eqref{eq:mcnemar}. Training and evaluation were run on a workstation
with an NVIDIA GeForce RTX 5060 GPU and an Intel Core Ultra 9 CPU.

One reproducibility caveat applies specifically to the MC-Dropout read-out. The seed fixes training
and data partitioning, but the $T=20$ stochastic forward passes are not re-seeded before each
separate evaluation call, so repeated read-outs of the same trained model draw fresh dropout masks
and can disagree on a handful of borderline cases. In practice the effect is about one image in
1{,}112 ($<0.1\%$): for example, ViT seed~42 test accuracy appears as 0.9631 in the main summary and
as 0.9640 in the separate McNemar re-evaluation. Consequently the paired accuracies in
Table~\ref{tab:mcnemar} come from a different inference pass than the discrimination and calibration
figures in Tables~\ref{tab:perf}--\ref{tab:perseed}, and may differ from them at the fourth decimal.
This does not affect any reported conclusion; the discrepancy is well within the bootstrap CIs and
never changes a class prediction that matters to the McNemar contingency counts. A fully
deterministic pipeline would fix the dropout RNG per evaluation call as well as per run.

\end{document}